\documentclass[11pt,a4paper]{article}
\usepackage{authblk}
\usepackage[hyperref]{conll-2019}
\usepackage{times}
\usepackage{latexsym}
\usepackage{multirow}
\usepackage{multicol}
\usepackage{amsmath}
\usepackage{graphicx}
\usepackage{amsfonts}
\usepackage{blkarray}
\usepackage[]{algorithm2e}
\usepackage{array}
\usepackage{booktabs}
\usepackage[utf8]{inputenc}
\usepackage[usestackEOL]{stackengine}

\usepackage{url}

\let\OLDthebibliography\thebibliography
\renewcommand\thebibliography[1]{
  \OLDthebibliography{#1}
  \setlength{\parskip}{0pt}
  \setlength{\itemsep}{0pt plus 0.1ex}
}

\aclfinalcopy % Uncomment this line for the final submission
\title{UniSent: Universal Adaptable Sentiment Lexica for 1000+ Languages}

\author[$\dag, \ddag, \diamond$]{Ehsaneddin Asgari}
\author[$\ddag$]{Fabienne Braune}
\author[$\diamond$]{Benjamin Roth}
\author[$\ddag$]{Christoph Ringlstetter}
\author[$\dag,\diamondsuit$]{Mohammad R.K. Mofrad}
\affil[$\dag$]{Department of Bioengineering, University of California, Berkeley, CA 94720, USA}
\affil[$\ddag$]{NLP Expert Center, Volkswagen Group Data:Lab, Munich, Germany}
\affil[$\diamond$]{Center for Information and Language Processing, Munich 80538, Germany}
\affil[$\diamondsuit$]{Molecular Biophysics and Integrated Bioimaging, Lawrence Berkeley National Lab, Berkeley, CA 94720, USA}

\date{}

\date{}

\begin{document}
\maketitle

\begin{abstract}
In this paper, we introduce \textit{UniSent} universal sentiment lexica for $1000+$ languages. Sentiment lexica are vital for sentiment analysis in absence of document-level annotations, a very common scenario for low-resource languages. To the best of our knowledge, \textit{UniSent} is the
largest sentiment resource to date in terms of
the number of covered languages, including many
low resource ones. In this work, we use a massively parallel Bible corpus to project sentiment information from English to other languages for sentiment analysis on  Twitter data.
We introduce a method called \textit{DomDrift} to mitigate the huge domain mismatch between Bible and Twitter by a confidence weighting scheme that uses domain-specific embeddings to compare the nearest neighbors for a candidate sentiment word in the source (Bible) and target (Twitter) domain.
We evaluate the quality of \textit{UniSent} in a subset of languages for which manually created ground truth was available, Macedonian, Czech, German, Spanish, and French. We show that the quality of \textit{UniSent} is comparable to manually created sentiment resources when it is used as the sentiment seed for the task of word sentiment prediction on top of embedding representations. In addition, we show that emoticon sentiments could be reliably predicted in the Twitter domain using only \textit{UniSent} and monolingual embeddings in German, Spanish, French, and Italian. With the publication of this
paper, we release the \textit{UniSent} sentiment lexica.  

\end{abstract}

%combining annotation projection,
%vocabulary expansion, and unsupervised domain
%adaptation based on source and targed embedding graph analysis

\section{Introduction}
Language technologies permeate our everyday life through web
search, translation, online shopping, email writing, spell
checking systems etc. The existence of such 
technologies highly depends on the existence
of the underlying computational linguistic
resources for a language. Computational linguistic resources such as
machine-readable lexica, part-of-speech-taggers and
dependency parsers are available for at most a few hundred
languages. This means that the majority of the 7000 languages
of the world are low-resource. This gap between advances in language technologies for English versus other languages endangers multilingualism in the digital age. Languages with lack of technological support (as a result of having limited resources) are less used over time and eventually get in danger of extinction. Many EU and US programs are designed to address
this issue~\cite{cieri2016selection}. The rationale of these projects is that even ``small'' languages are important for the preservation of the common
heritage of humankind and cultural diversity which benefits everybody. In addition, certain low-resource
languages can be also politically and economically important.\\
\indent Large amount of text available on the web and applications in marketing~\cite{bollen2011twitter}, social science~\cite{hopkins2010method}, political science~\cite{wang2012system,wong2016quantifying} motivates sentiment analysis of news, blogs, social networks, reviews, opinions, and recommendations. However, sentiment analysis requires either word or document level sentiment annotations. Typically, these are available only for a limited number of languages, preventing accurate sentiment classification in low resource setups. In these scenarios sentiment lexica are important game changers because in many cases end-to-end sentiment classification is not feasible due to a lack of document level annotations~\cite{jurafsky2014speech}. 

Recently, embedding-based approaches for supervised or semi-supervised word sentiment inference became popular allowing for lexicon vocabulary expansion and implicit domain adaptation~\cite{rothe-etal-2016-ultradense, hamilton2016inducing}. Although using the embedding space as representation in word sentiment classification sufficiently addresses domain adaptation in many cases, it can be improved in situations where the domain shift for some words from the lexicon seeds in the source vocabulary (e.g Bible\footnote{Massively parallel corpora mainly exist in the Bible domain and for a smaller text size for the universal declaration of human rights~\cite{emerson2014seedling} .}) to the target vocabulary (e.g. Twitter) is large. For instance, in Biblical texts, the Spanish word \textit{sensual} has the connotation of \textit{sin} which has a negative polarity. But in the Twitter domain, the same word is associated with \textit{sexy}, which has a positive polarity. In cases where the classifier using the embedding space fails to capture this shift, enhancing the model via mitigation of the domain mismatches is required.

\paragraph{Contributions:} We release the first sentiment lexicon covering 1000+ languages and achieving macro-F1 over 0.75 on word sentiment prediction for most evaluated languages, meaning that we enable sentiment analysis in many low resource languages. The creation of \textit{UniSent} requires only a sentiment lexicon in one language (e.g. English) and a small, but massively parallel corpus in a specific domain.
We evaluate \textit{UniSent} for word sentiment classification of Macedonian, Czech, German, Spanish, French against manually assigned sentiment polarities and show that its quality is comparable to the use of manually created resources, which is a great evidence that \textit{UniSent} works well also for low-resource languages where we do not have resources for evaluation. Secondly, we evaluated UniSent w.r.t. the classification of emoticon sentiments in the Twitter domain, where macro-F1 of 0.79, 0.76, 0.74, and 0.76 were obtained for German, Italian, French, and Spanish respectively.  

\indent To ensure the usability of our lexica for any new domain, we propose \textit{DomDrift}, a method requiring only a pretrained embedding space in the target domain, which is relatively a cheap resource to obtain. By comparing the source and target embedding graphs~\textit{DomDrift} quantifies the semantic changes of words in the sentiment lexicon in the new domain. This measure can hence be used to weight words in the sentiment lexicon for downstream supervised or semi-supervised sentiment analysis models. We show that on top of implicit domain adaptation, using target domain embeddings, the incorporation of domain drift scores improves sentiment classification for French, Spanish, and Macedonian. 

\section*{Related Work}
Several research efforts tackled the automatic creation of sentiment lexica for a multitude of languages, but these efforts resulted in the creation of resources for at most 136 languages~\cite{chen-skiena-2014-building} or in lexicon covering a very specific low-resource language~\cite{afli2017sentiment,darwich2017minimally}. Moreover, these approaches heavily rely on linguistic resources, such as WordNet or fully trained machine translation systems, which limit them to the languages where these are available. An alternative to our approach in lexicon creation for sentiment is using minimal bilingual supervision \cite{hangya2018, barnes2018, barnes2018-2} to create document-level annotations for end-to-end sentiment classifications of documents. Later approaches only work in an end-to-end fashion and do not allow to directly create sentiment lexica.

The annotation projection to create  \textit{UniSent} sentiment lexica is inspired by \textit{SuperPivot} introduced in  \cite{asgari2017pastPresentFuture}
for the typological analysis of tense in 1000 languages. \newcite{Agic2016} also use massively parallel corpora to project POS tags and dependency relations across languages. In contrast to these studies, here we perform parallel projection on sentiment information for resource creation and not for typological analysis. In addition, we propose a method called \textit{DomDrift} to mitigate the huge domain mismatch between Bible and target domain via an embedding-based confidence weighting scheme.  

\section{Methods}
In the next sections, we describe  (i) the main resources required for \textit{UniSent} and (ii) the steps  of its creation and adaptation to new domains. The overview of these steps is also depicted in Figure~\ref{fig:overview}.

\subsection{UniSent Required Resources}
\textbf{Super-parallel corpus:} The dataset we will work with is the Parallel Bible Corpus (PBC).
PBC consists of translations of the New Testament in 1242 languages covering an order of magnitude more languages than any other parallel corpus currently in use in natural language processing research~\cite{mayer2014creating}.\\

\noindent\textbf{Initial sentiment seeds:} We use a high-quality English sentiment lexicon called WKWSCI \cite{khoo2018lexicon} as a resource to be projected on other languages.

\subsection{UniSent and DomDrift}
\begin{figure*}[ht]
\centering
  \includegraphics[trim=0 0 0 0, width=1.01\textwidth,clip]{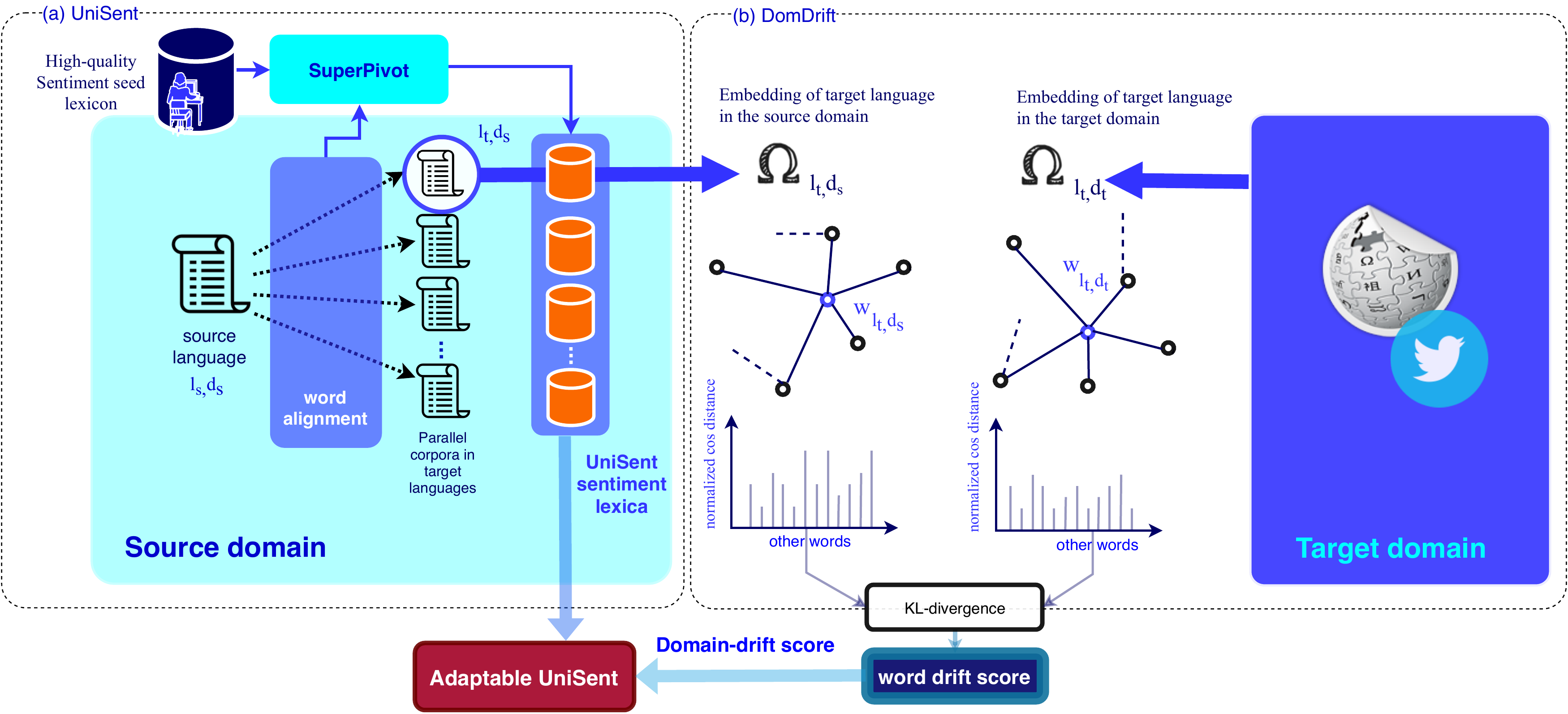}
  \caption{\label{fig:overview} The overview of universal adaptable sentiment lexica. The approach can be divided into two main steps: (a) UniSent creation using SuperPivot method, (b) DomDrift for measurement of word domain drifts.}
\end{figure*}

Our contributions are two-fold (i) creation of \textit{UniSent} using a cross-lingual projection of sentiment polarities, which needs to be done only once (ii) introducing \textit{DomDrift} a novel method for adapting \textit{UniSent} to any newly observed domain by measuring the domain-drift of words in the new domain. The first part needs a sentiment lexicon in one language (here WKWSCI for English) as well as a massively parallel corpus (PBC). For the second part, only a pre-trained embedding space in the target domain is required. In the next sections, we illustrate our method by creating \textit{UniSent} for one example language (for better readability). The described steps are however repeated for each of the  1000+ languages composing \textit{UniSent}. The steps are detailed next and illustrated in Figure~\ref{fig:overview}.

\noindent\paragraph{(i) UniSent creation using cross-lingual projection of sentiment polarities:}
\label{para:projection}
We project sentiment polarities from English (source language) to a target language in the parallel corpus using \textit{SuperPivot}~\cite{asgari2017pastPresentFuture}. This method projects annotations across 1000+ languages via an alignment graph generated using \texttt{FastAlign}~\cite{dyer2013simple} on the PBC corpus. In the \texttt{FastAlign} word alignment pairs $(w_{source},w_{target})$, we replace the source words with their sentiment labels from WKWSCI (where available). Subsequently, we search for the words in the target language that are highly correlated with each of the sentiment labels (positive or negative). We use FDR corrected two-sided $\chi^2$~\cite{casella2002statistical} score to find these sentiment seeds. We denote the vocabulary of the target language in the parallel corpus by $V_{l_t,d_s}$, where $l_t$ is the target language and $d_s$ the source domain (here the domain of the parallel corpus, i.e. biblical domain).\footnote{We call this domain \textit{source domain} because it will be adapted to a target domain in the subsequent steps.}
This first step generates, in each target language, pairs $(w_{l_t,d_s}, y)$, where $w_{l_t,d_s} \in V_{l_t,d_s}$ is a word in the target language and source domain and $y$ is a highly correlated sentiment annotation with $w_{l_t,d_s}$. Vocabulary $V_{l_t,d_s}$ is limited to the words in the parallel corpus~\footnote{Because we use this corpus for the cross-lingual projection}. Because most existing super-parallel corpora are from the Bible~\cite{Mayer2014}, $V_{l_t,d_s}$ besides of being limited in size has the major drawback of originating from a very specific domain. To overcome this limitation, we define a method to measure domain drifts. In addition, we leverage word embeddings as used in~\cite{rothe-etal-2016-ultradense} to propagate the annotations $y$ to words of a larger vocabulary than $V_{l_t,d_s}$. The overview of UniSent creation is depicted in Figure~\ref{fig:overview}.a. The \textit{UniSent} lexica and a complete list of 1242 unique languages\footnote{We consider two languages different if they have different ISO 639-3 codes} covered by \textit{UniSent} along with their language family information are provided in the supplementary material.

\noindent\paragraph{(ii) DomDrift: unsupervised measurement of domain drift:}
\label{para:adaptation}
Word embeddings trained with an unsupervised language modeling objective (e.g., skip-gram) are known to preserve the syntactic and the semantic word similarities in the embedding space~\cite{NIPS2013_5021,pennington2014glove}. Domain changes will certainly impact the neighborhoods in the embedding space. Thus a comparison of words relative distances in two embedding spaces can be used to measure their degree of domain shift~\cite{kulkarni2015statistically,asgari2016comparing}. The main purpose of \textit{DomDrift} is to identify words in the sentiment lexicon having a domain shift in the target domain by comparison of their neighbors in the embedding spaces.

\noindent\textit{DomDrift} quantifies the domain drift of a given word in the sentiment lexicon regardless of its label, only by comparison of neighbors in the source and target domains' embedding spaces $\Omega_{l_t,d_s}: V_{l_t,d_s} \xrightarrow{}$ $\mathbb{R}^{h_s}$,  and $\Omega_{l_t,d_t}: V_{l_t,d_t} \xrightarrow{} \mathbb{R}^{h_t}$, where $h_s$, $h_t$ are the sizes of the source and target embedding spaces. \textit{DomDrift}
quantifies word domain drift as follows:
\begin{enumerate}
    \item[(i)] In each embedding space, we compute, for each word $w_{l_t,d_x}$ in the UniSent lexicon, the distance distribution $P(w_{l_t,d_x},\boldsymbol{w}_{\Omega_{s,t}})$ of word $w_{l_t,d_x}$ with all other words in intersection of source and target embedding spaces, i.e. $\forall~ w_j \in \boldsymbol{w}_{\Omega_{s,t}}$. For this, we take the $l1$ normalized cosine distance of the representation of $w_{l_t,d_x}$ with all other words in the embedding space. This distribution can be regarded as word profile in the domain x (source or target):
   \[
    P_{i}(w_{l_t,d_x},\boldsymbol{w}_{\Omega_{s,t}})= \frac{1-cos(\overrightarrow{w_{l_t,d_x}}, \overrightarrow{w_i})}{\sum_{j}  [1-cos(\overrightarrow{w_{l_t,d_x}}, \overrightarrow{w_j})]},
   \]
    where $w_i, w_j \in \Omega_{s,t}$ and $\overrightarrow{w_k}$ is the embedding representation of word $w_k \in \Omega_{s,t}$ in domain x (source or target). We use $\boldsymbol{w}_{\Omega_{s,t}}$ so that the word profiles in the source and target domains are comparable, i.e. they have the same elements.
    \item[(ii)] We compute, for each word $w_{l_t,d_x}$ a shift weight between vocabularies $V_{l_t,d_s}$ and $V_{l_t,d_t}$ of source and target spaces. This is done by comparing, for each word in the UniSent lexica, its profile in the source $P(w_{l_t,d_s}, \boldsymbol{w}_{\Omega_{s,t}})$ and target domain $P(w_{l_t,d_t}, \boldsymbol{w}_{\Omega_{s,t}})$ using the Kullback-Leibler divergence.
\end{enumerate}

More formally for a word $w'$ its domain drift ($\lambda_{w'}$) between the source and the target domains can be calculated as follows.

\[\lambda_{w'}= D_{\mathrm{KL}}(P({w'}_{l_t,d_s},\boldsymbol{w}_{\Omega_{s,t}})\|P({w'}_{l_t,d_t},\boldsymbol{w}_{\Omega_{s,t}})) \]

\begin{figure*}
\centering
  \includegraphics[width=1.1\textwidth,clip]{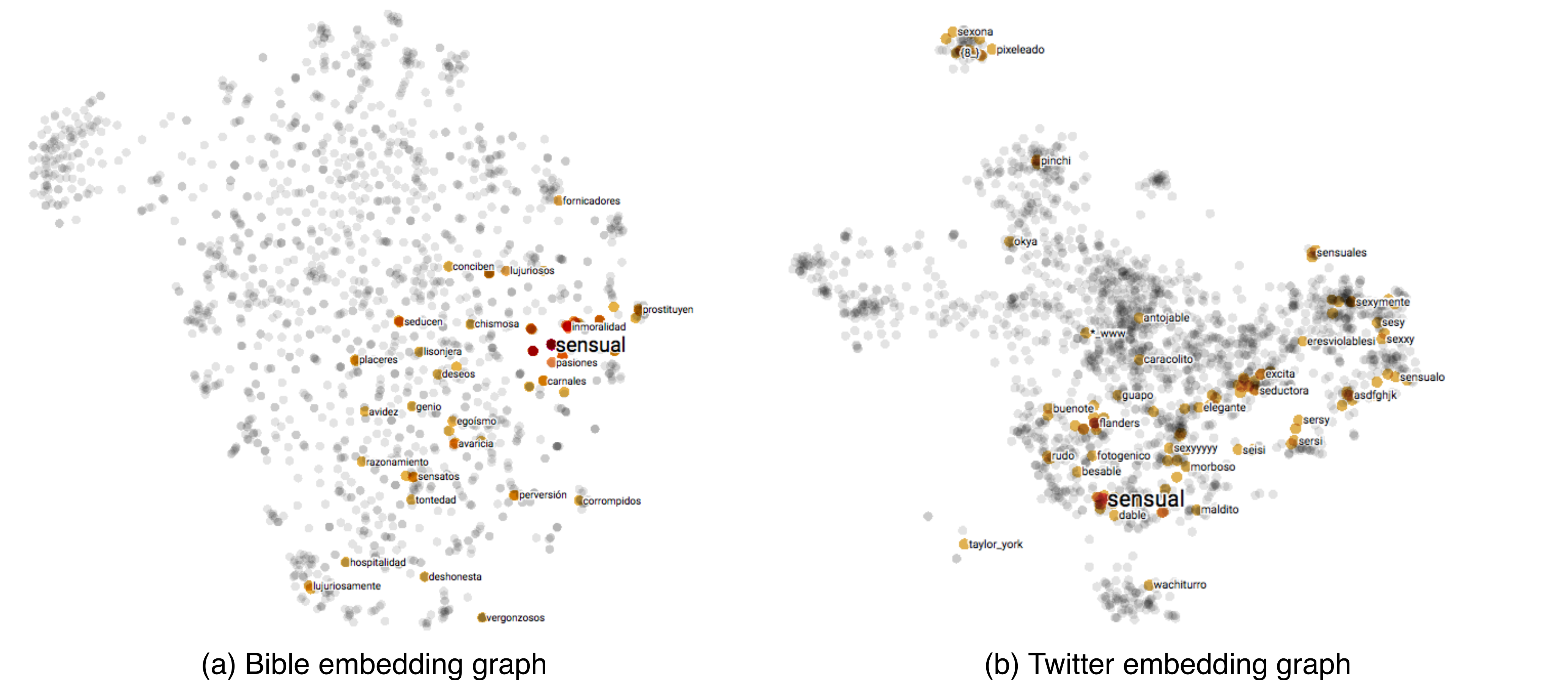}
  \caption{\label{fig:domainShift} Neighbors of the word 'sensual' in Spanish, in the bible embedding graph (a) and the twitter embedding graph (b). Our unsupervised drift weighting method found this word in Spanish to be the most changing word from bible context to the twitter context. Looking more closely at the neighbors, the word sensual in the biblical context has been associated with negative sentiment of sins. However, in the twitter domain, it has a positive sentiment. This example shows how our unsupervised method can improve the quality of sentiment lexica.}
\end{figure*}

The steps of DomDrift are illustrated in Figure~\ref{fig:overview}.b. As also depicted in the figure, the calculated weights enhance the universal sentiment lexica resulting in a final adaptable version (Adaptable UniSent), e.g. the weights will be used to reduce the influence of huge domain mismatches in a confidence weighting scheme. In Figure \ref{fig:domainShift} we illustrate an example of domain drift and explain the workings of our weighting method in $\S$\ref{evaluation}. Once we computed our shift weights,  they can be used in any semi-supervised or supervised approach (e.g., sample weights in the logistic regression model).\\

\noindent\textbf{Source embedding $\Omega_{l_t,d_s}$}:
In order to generate $\Omega_{l_t,d_s}$, the only necessary resource is the monolingual text of PBC in the target language (source domain). For embedding creation, we use \texttt{fasttext}~\cite{bojanowski2017} which leverages subword information within the skip-gram architecture.\\

\noindent\textbf{Target embedding $\Omega_{l_t,d_t}$}: For generation of $\Omega_{l_t,d_t}$, we require a monolingual text collection in the target domain to train the embedding space. An alternative is to use pretrained embeddings in the domain of interest (e.g. Twitter or News). In particular, in our experiments, we use skip-gram embeddings, pretrained on Wikipedia for French, Macedonian, Spanish, Czech, and German \cite{conneau2017word}, as well as German, Italian, French, and Spanish monolingual pretrained embeddings on Twitter, provided  by~\cite{deriu2017leveraging,cieliebak2017twitter}.

\subsection{Word sentiment classification for the evaluation of sentiment lexica}
\label{para:propagation}
In order to evaluate \textit{UniSent}, we use it as seed lexicon for word sentiment classification on top of embedding features. Different methods can be used to predict the sentiment of words in the target domain using embedding spaces. These include supervised methods e.g., UltraDense~\cite{rothe-etal-2016-ultradense}, linear classifiers and regressors (e.g. SVM, SVR, and logistic regression) or semi-supervised methods, e.g. SentProp~\cite{hamilton2016inducing}. In this paper, we use logistic regression for the classification model. Using the language model based embedding space as the representation has two major benefits:  First, the semantic continuity of the embedding space allows for propagation of sentiment labels to a larger vocabulary. This enables the annotation of further word pairs $(w_{l_t,d_t}, \hat{y})$, where $w_{l_t,d_t} \in V_{l_t,d_t}$ is a vocabulary of arbitrary size in the target domain. Secondly, using embeddings trained in a specific domain result in the implicit incorporation of semantic structures specific to the target domain (which are reflected in the embedding space), i.e. an implicit domain adaptation.

\noindent\textbf{UniSent versus confidence weighted UniSent:} For the  word sentiment classification we train a logistic regression classifier with the annotated pairs of $(w_{l_t,d_t}, {y})$ represented in the target embedding space. In order to incorporate the drift weights, we treat the weights coming from \textit{DomDrift} as sample weights in the logistic regression model, i.e. $(w_{l_t,d_t}, {y}, s_w)$'s are the training instances to the logistic regression classifier, where $s_w = \frac{1}{\lambda_w}$, is the seed weight calculated as the inverse of \textit{DomDrift} score. Our evaluation in section~\S\ref{evaluation} shows that this (simple) method is very effective and creates accurate resources. In the hyper-parameter tuning for logistic regression we also fine tune the exponent of this weight.\\

\section{Experiments and Evaluation}
\label{experiments}
\subsection{Experimental Setup}
\label{setup}
\paragraph{Select Gold Standard Data}
As gold standard sentiment lexica for the evaluation of UniSent, we select manually created lexica in Czech~\cite{veselovska2013czech}, German~\cite{waltinger2010germanpolarityclues}, French~\cite{abdaoui2014feel}, Macedonian~\cite{jovanoski2016impact}, and Spanish~\cite{perez2012learning}. These lexica contain general domain words (as opposed to Twitter or Bible). As gold standard for Twitter we use the emoticon dataset in  ~\cite{wiebe2005annotating,hogenboom2013exploiting} and perform emoticon sentiment prediction for different languages.

\subsection{Train-test split}
\label{sec:testSets}
\begin{figure}[ht]
\centering
  \includegraphics[width=0.5\textwidth,clip]{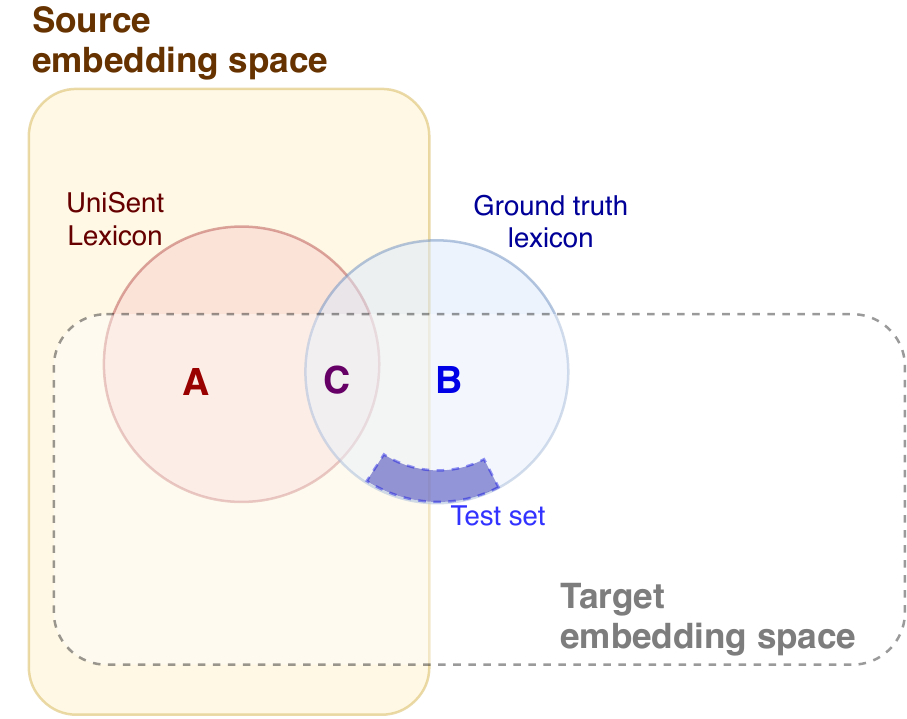}
  \caption{\label{fig:datasplit} Data split used in the experimental setup of UniSent evaluation: Set (C) is the intersection of the target embedding space words (Wikipedia or Twitter) and the UniSent lexicon as well as the manually created lexicon. Set (A) is the intersection of the target embedding space words and the UniSent lexicon, excluding set (C). Set (B) is the intersection of the target embedding space words and the manually created lexicon, excluding set (C).}
\end{figure}

In order to evaluate the UniSent, here we create train-test split for training and testing the seeds created in the projection step (see Section $\S$ \ref{para:projection}). We first split \textit{UniSent} and our gold standard lexica as illustrated in Figure~\ref{fig:datasplit}. In order to design a fair evaluation, we form our training and test sets as follows:

\noindent\textbf{(i) UniSent-Train-Lexicon:} For the evaluation of the UniSent, we use words in \textit{UniSent} as sentiment seeds for training in the target domain; for this purpose, we use words $w \in A \cup C$ (Figure~\ref{fig:datasplit}).

\noindent\textbf{(ii) Manual-Train-Lexicon:} In order to obtain an upper bound for the UniSent performance, we compare the use of UniSent-Train-Lexicon against the use of words in the gold standard lexicon as sentiment seeds for the training in the target domain. For this purpose, we use words $w \in B \cup C$ (Figure~\ref{fig:datasplit}).

\noindent\textbf{(iii) Test-Lexicon:} we randomly exclude a set of words in the \{\textbf{Manual-Train-Lexicon} $\cup B$\}. In the selection of the sampling size, we make sure that $UniSent-Train-Lexicon$ and $Manual-Train-Lexicon$ would contain approximately the same number of words (Figure~\ref{fig:datasplit}).

\subsection{Evaluation}
\label{eval}

As discussed in~$\S$\ref{para:projection}, we use the (manually created) English sentiment lexicon (WKWSCI) in \cite{khoo2018lexicon} as a resource to be projected to over 1000+ languages. We project positive and negative sentiments to create positive and negative sentiment lexica for each language.

Our evaluation of this work is two-fold. On the one hand, we evaluate the overall quality of \textit{UniSent} by comparing it against our manually created gold standard datasets in the Wikipedia domain. Second, we investigate the influence of~\textit{DomDrift} w.r.t.  the adaptation of UniSent for Wikipedia and Twitter domains.

\noindent\textbf{(i) Evaluation of UniSent vs. manually created lexica:} We compare the application of Unisent for the word sentiment classification task against the manually created lexica in the following cases: (i) choice of the most frequent sentiment, (ii) use of manually created lexicon as sentiment seeds. We use the train and test seed lexica as discussed in~\ref{sec:testSets} for training and testing of the logistic regression on top of target embedding features. We also include the sentiment classification results using the confidence weighted version of \textit{UniSent}, where the shift between the vocabularies of the Bible and Wikipedia are calculated by \textit{DomDrift} and are used as sample weights in the logistic regression model.

\noindent\textbf{(ii) Comparison of UniSent vs. confidence weighted UniSent in the Twitter domain for emoticon prediction:} To show that our adaptation method also works well on domains like Twitter, we propose a second evaluation in which we use \textit{UniSent} together with \textit{DomDrift} to predict the sentiment of emoticons in Twitter. Since emoticons are almost language independent, we could use the same resource for the evaluation of German, Italian, French, and Spanish, where their monolingual pretrained embeddings are available for these languages~\cite{deriu2017leveraging,cieliebak2017twitter}. In the adaptation step, we compute the shift between the vocabularies of Bible and Twitter. We use the \textit{UniSent} seeds for training a logistic regression model on Twitter embedding and evaluate the classifier for the Emoticon sentiment prediction. We perform this evaluation for German, Italian, French, and Spanish, where Twitter pretrained embedding is available.

\begin{table*}[ht]
\centering
 \caption{\label{tab:sentiment_res_1} We evaluate \textit{UniSent} against the gold standard datasets in Czech, German, French, Macedonian, and Spanish. The two last columns report the accuracy and macro-F1 (averaged F1 over positive and negative classes) of \textit{Unisent} before and after the application of the drift weighting step. The two first columns report the performance of the baseline and manually created lexicon. Note that the baseline is constantly considering the majority label.}

\resizebox{2\columnwidth}{!}{\begin{tabular}{c|c|cc|cc|cc}
\toprule
\small{Language} & \small{Freq. sentiment baseline}  & \multicolumn{2}{c}{\Longstack{\small{Manual Lexicon}\\target-language-specific}} & \multicolumn{2}{c}{\Longstack{\small{UniSent Lexicon}\\(projection)}} & \multicolumn{2}{c}{\Longstack{\small{Confidence Weighted UniSent Lexicon}\\(projection)}}\\
 & \small{acc} & \small{acc} & \small{macro-F1} & \small{acc} & \small{macro-F1}  & \small{acc} & \small{macro-F1} \\\midrule\midrule
French	& 0.62 &    0.84 & 	0.83 & 0.73& 	0.72& 	0.74& 	0.74\\
Macedonian	& 0.70 & 	0.86 & 	0.84& 	0.80& 	0.77& 	0.81& 	0.78\\
Spanish	& 0.64 & 	0.82 & 	0.80& 	0.78& 	0.76& 	0.80& 	0.77\\
Czech	& 0.62 & 	0.87 &  0.87& 	0.82& 	0.81& 	0.79& 	0.78\\
German	& 0.52 &    0.87 &  0.87& 	0.82& 	0.81& 	0.81& 	0.80\\
\bottomrule
\end{tabular}}
\end{table*}

\begin{table*}[ht]
\centering
 \caption{\label{tab:sentiment_res_2}
 We evaluate \textit{UniSent} using twitter emoticon dataset. We use monolingual Twitter embeddings in German, Italian, French, and Spanish. The two last columns report the accuracy and macro-F1 (averaged F1 over positive and negative classes) of \textit{Unisent} before and after the application of the drift weighting step.}
\resizebox{1.7\columnwidth}{!}{\begin{tabular}{c|c|cc|cc}
\toprule
\small{Language} & \small{Freq. sentiment baseline} & \multicolumn{2}{c}{\Longstack{\small{UniSent Lexicon}\\(projection)}} & \multicolumn{2}{c}{\Longstack{\small{Confidence Weighted UniSent Lexicon}\\(projection)}}\\
 & \small{acc} &  \small{acc} & \small{macro-F1}  & \small{acc} & \small{macro-F1} \\\midrule\midrule
French	& 0.62 &    0.73 & 	0.73 & 0.75 & 	0.74\\
Spanish	& 0.62 & 	0.73 & 	0.73 & 	0.76 & 	0.76\\
German	& 0.62 &    0.80 &  0.79 & 	0.80& 	0.79\\
Italian	& 0.62 & 	0.76 & 	0.76 & 	0.75 & 	0.75\\
\bottomrule
\end{tabular}}
\end{table*}
%\vspace{-0.5cm}
\subsection{Results}
\label{evaluation}
The evaluation results are reported in Tables \ref{tab:sentiment_res_1} and \ref{tab:sentiment_res_2}. Table \ref{tab:sentiment_res_1} compares \textit{UniSent} and its confidence weighted version to the manually created lexica in Czech, German, French, Macedonian, and Spanish as well as a naive baseline of choosing the most frequent sentiment. For all evaluated languages, accuracy as well as macro-F1 are close to 0.8, showing that \textit{UniSent} is a high-quality resource performing close enough to manually created seeds and clearly better than the most frequent sentiment baseline. Since in this evaluation the presented languages did not have any further advantage than having a manually created lexicon for evaluation, we can assume that UniSent would work within the same range of accuracy for any of the low-resource languages as long as a monolingual embedding (which is also cheap to obtain for low-resource languages) can be available for the target domain. Our drift weighting method brings gains in several languages: French, Macedonian, and Spanish.

In Table \ref{tab:sentiment_res_2} we compare the quality of \textit{UniSent} in prediction of the gold standard emoticon sentiments in the Twitter domain. The results show that (i)  \textit{UniSent} clearly outperforms the baseline of the most frequent sentiment label and (ii) our domain adaptation technique brings small improvements for French and Spanish.

In order to illustrate the function of \textit{DomDrift} we visualized the embedding space of biblical domain and twitter domain for the word achieving the highest drift score in Spanish, i.e., word \textit{sensual} (Figure~\ref{fig:domainShift}). The neighborhood of this word in both domains is shown in the figure. In Biblical texts, this word has the connotation of \textit{sin} which has a negative polarity. But in the Twitter domain, the same word is associated with \textit{sexy}, which has a positive polarity. This example shows that for certain pairs of domains and languages use of \textit{DomDrift} weights in the use of sentiment lexicon can improve the performance of sentiment analysis.

\section{Discussion and Conclusion}

In this work, we introduced \textit{UniSent}  universal sentiment lexica for $1000+$ languages, which is to the best of our knowledge, the
largest sentiment resource to date in terms of the number of covered languages, including many low resource ones. Although \textit{UniSent} is created based on a specific domain (bible), our evaluation of \textit{UniSent} on Czech, German, French, Macedonian, and Spanish showed that it can achieve macro-F1 scores $\approx$0.8 in word sentiment classification, which is comparable to the use of manually annotated resources. Given that many of covered languages in \textit{UniSent} are very low-resource, \textit{UniSent} can be regarded among the only computational linguistic resources available for those low-resource languages making sentiment analysis possible in such low-resource setups. In addition, through accurate prediction of Twitter emoticon sentiment for German, Italian, French, and Spanish, we showed that \textit{UniSent} can be even used in a very different target domain and still performs quite well in sentiment analysis.

Furthermore, we proposed \textit{DomDrift}, a method to quantify domain drift for words in the UniSent given an embedding space in the target domain. \textit{DomDrift} compares the neighborhood of the word in the embedding spaces of source and target domains. Incorporation of \textit{DomDrift} scores in the use of \textit{UniSent} for sentiment classification outperformed vanilla \textit{UniSent} in French, Spanish, and Macedonian in the Wikipedia domain and French and Spanish in the twitter domain.  Not further improving the results for German, Czech and Italian languages might be because of the sufficiency of the target embedding usage for domain adaptation~\cite{jurafsky2014speech} in those. On the other hand, the fact that Spanish and French performances improved on both Wikipedia and Twitter domains when \textit{DomDrift} is used, might show that the necessity of \textit{DomDrift} can be related to certain property of the target language, which can be further explored as future work.

\bibliographystyle{acl_natbib}

\end{document}